# A Geometric Approach to Mapping Bitext Correspondence

I. Dan Melamed

Dept. of Computer and Information Science
University of Pennsylvania
Philadelphia, PA, 19104, U.S.A.
melamed@unagi.cis.upenn.edu

## Abstract

The first step in most corpus-based multilingual NLP work is to construct a detailed map of the correspondence between a text and its translation. Several automatic methods for this task have been proposed in recent years. Yet even the best of these methods can err by several typeset pages. The Smooth Injective Map Recognizer (SIMR) is a new bitext mapping algorithm. SIMR's errors are smaller than those of the previous front-runner by more than a factor of 4. Its robustness has enabled new commercial-quality applications. The greedy nature of the algorithm makes it independent of memory resources. Unlike other bitext mapping algorithms, SIMR allows crossing correspondences to account for word order differences. Its output can be converted quickly and easily into a sentence alignment. SIMR's output has been used to align over 200 megabytes of the Canadian Hansards for publication by the Linguistic Data Consortium.

## 1 Introduction

The first step in most corpus-based multilingual NLP work is to construct a detailed map of the correspondence between a text and its translation (a **bitext map**). Several automatic methods have been proposed for this task in recent years. However, most of these methods address only the sub-problem of alignment [CRW89, BLM91, G&C91, D&S92, SFI92, K&R93, Wu94]. Alignment algorithms assume the availability of text unit boundary information and their output has less expressive power than a general bitext map. The only published solution to the more difficult general bitext mapping problem [Chu93] can err by several typeset pages. Such frailty can expose lexicographers and terminologists to spurious concordances, feed noisy training data into statistical translation models, and degrade the performance of corpus-based machine translation. Some multilingual NLP tasks, such as automatic validation of terminological consistency [Mac95] and automatic detection of omissions in translations (implemented for the first time in [Mel96]), have been technologically impossible until now, because they are highly sensitive to large errors in the bitext map.

The Smooth Injective Map Recognizer (SIMR) is a greedy algorithm for mapping bitext correspondence. SIMR borrows several insights from previous work. Like Gale & Church [G&C91] and Brown et al. [BLM91], SIMR relies on the high correlation between the lengths of mutual translations. Like char_align [Chu93], SIMR infers bitext maps from likely points of correspondence between the two texts, points that are plotted in a two-dimensional space of possibilities. Unlike previous methods, SIMR searches for only a handful of points of correspondence at a time.

Each set of correspondence points is found in two steps. First, SIMR generates a number of possible points of correspondence between the two texts, as described in Section 3.1. Second, SIMR selects those points whose geometric arrangement most resembles the typical arrangement of *true* points of correspondence. This selection involves localized pattern recognition heuristics, which Section 3.2 refers to collectively as the **chain recognition heuristic**. SIMR then interpolates between successive selected points to produce a bitext map, as described in Section 3.3.

## 2 Definitions

Several key terms will help to explain SIMR. First, a **bitext** [Har88] comprises two versions of a text, such as a text in two different languages. Translators create a bitext each time they translate a text. Second, each bitext defines a rectan-gular **bitext space**, as illustrated in Figure 1. The width and height of the rectangle are the lengths of the two component texts, in characters. The lower left corner of the rectangle is the **origin** of the bitext space and represents the two texts' beginnings. The upper right corner is the **terminus** and represents the texts' ends. The line between the origin and the terminus is the **main diagonal**. The slope of the main diagonal is the **bitext slope**.

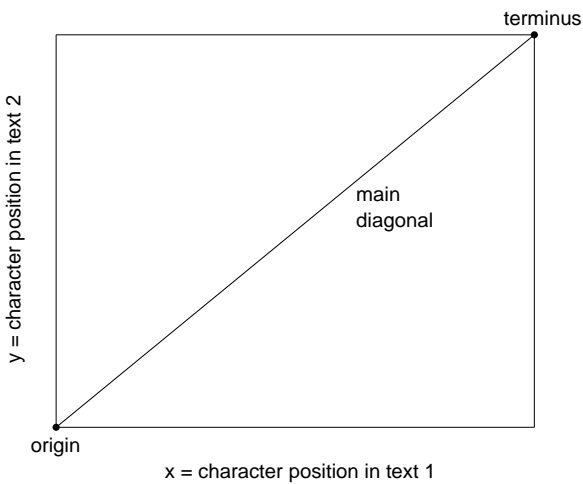

Figure 1: *a bitext space*

Each bitext space contains a number of **true points of correspondence (TPCs)**, other than the origin and the terminus. For example, if a token at position $p$ on the x-axis and a token at position $q$ on the y-axis are translations of each other, then the coordinate $(p, q)$ in the bitext space is a TPC[1]. TPCs also exist at corresponding boundaries of text units such as sentences, paragraphs, and chapters. Groups of TPCs with a roughly linear arrangement in the bitext space are called **chains**.

**Bitext maps** are bijective functions in bitext spaces. For each bitext, the **true bitext map (TBM)** is the shortest bitext map that runs through all the TPCs. The purpose of a **bitext mapping algorithm** is to produce bitext maps that are the best possible approximations of each bitext's TBM.

## 3 SIMR

Most of SIMR's effort is spent searching for TPCs, one short chain at a time. The search for each chain begins in a small rectangular region of the bitext space, whose dimensions are proportional to those of the whole bitext space. Within this search rectangle, the search alternates between a generation phase and a recognition phase, which are described in more detail in Sections 3.1 and 3.2. In the generation phase, SIMR generates all the points of correspondence that satisfy the supplied matching predicate (explained below). In the recognition phase, SIMR calls the chain recognition heuristic to search for suitable chains among the generated points. If no suitable chains are found, the search rectangle is proportionally expanded up and to the right and the generation-recognition cycle is repeated. The rectangle keeps expanding until at least one acceptable chain is found. If more than one chain is found, SIMR accepts the chain whose points are least dispersed around its least-squares line. Then, SIMR selects another region of the bitext space to search for the next chain.

---

[1]Since distances in the bitext space are measured in characters, the position of a token is defined as the mean position of its characters.

SIMR employs a simple heuristic to select regions of the bitext space to search. To a first approximation, TBMs are monotonically increasing functions. This means that if SIMR accepts a chain, it should look for others either above and to the right or below and to the left of the one it has just located. All SIMR needs is a place to start the trace, and a good place to start is at the beginning. The origin of the bitext space is always a TPC. So, the first search rectangle is anchored at the origin. Subsequent search rectangles are anchored at the top right corner of the previously found chain, as shown in Figure 2.

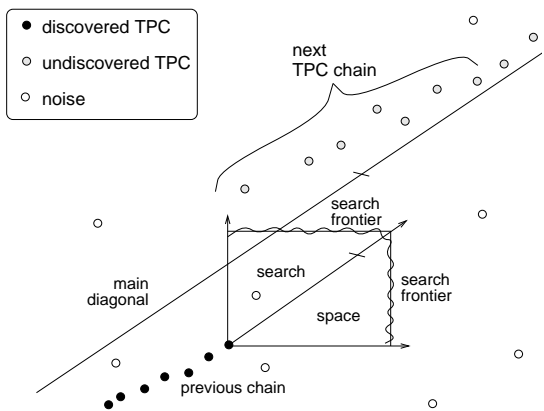

Figure 2: *SIMR's "expanding rectangle" search strategy. The search rectangle is anchored at the top right corner of the previously found chain. Its diagonal remains parallel to the main diagonal.*

The expanding-rectangle search strategy makes SIMR robust in the face of TBM discontinuities. Figure 2 shows a segment of the TBM trace that contains a vertical gap (an omission in the text on the x-axis). As the search rectangle grows, it will eventually pick up the TBM's trail, even if the discontinuity is quite large [Mel96]. Section 3.8 explains why SIMR will not be led astray by false points of correspondence.

### 3.1 Point Generation

A **matching predicate** is a heuristic for guessing whether a given point in the bitext space is a TPC. I have considered only token-based matching predicates, which can only return TRUE for a point $(x, y)$ if $x$ is the position of a token e on the x-axis and $y$ is the position of a token f on the y-axis. For each such point, the matching predicate must decide whether the e and f are likely to be mutual translations.

Various knowledge sources can be brought to bear on the decision. The most universal knowledge source is a translation lexicon. Translation lexicons can be extracted from machine-readable bilingual dictionaries (MRBDs), in the rare cases where MRBDs are available. In other cases, they can be induced automatically using any of several existing methods [DCG93, F&C94, Mel95]. Since the matching predicate does not require perfect accuracy, the induced lexicons need not be perfect. When a large translation lexicon is not available, a small hand-constructed translation lexicon for the key terms in a given bitext may suffice to produce a rough map for that bitext.

If the languages involved have similar alphabets, then it may be possible to construct a matching predicate with very little effort, using the method of cognates. Cognates are words with a common etymology and a similar meaning in different languages. The etymological similarity is often reflected in the words' orthography and/or pronunciation. Languages that are closely related will often share a large number of cognates. For example, in the non-technical Canadian Hansards (parliamentary debate transcripts available in English and French), cognates can be found for roughly one quarter of all text tokens [Mel95]. A cognate-based matching predicate will generate more points for more similar language pairs, and for text genres where more word borrowing occurs, such as technical texts. For English and French, such a matching predicate can generate enough points in the bitext space to obviate the need for a translation lexicon.

Phonetic cognates can be used to map between language pairs with dissimilar alphabets, even when the languages are not closely related. When language L1 borrows a word from language L2, the word is usually written in L1 similarly to the way it sounds in L2. Thus, French and Russian /pɔrtmɔne/ are cognates, as are English /sIstəm/ and Japanese /šisutemu/. For many languages, it is not difficult to construct an approximate mapping from the orthography to its underlying phonological form. Given such a mapping for L1 and L2, it is possible to identify cognates despite incomparable orthographies.

SIMR was tested on French and English with two different matching predicates. The first matching predicate relies on orthographic cognates and a stop-list of closed-class words for both languages. SIMR judges the cognateness of each token pair by their Longest Common Subsequence Ratio (LCSR). The LCSR of a token pair is the number of characters that appear in the same order in both tokens divided by the length of the longer token [Mel95]. The common characters need not be contiguous. The matching predicate considers a token pair cognates if their LCSR exceeds a certain threshold. The LCSR threshold was optimized together with SIMR's other parameters, as described in Section 3.7. The stop-list of closed-class words made the matching predicate more accurate, because closed-class words are unlikely to have cognates. On the contrary, they often produce spurious matches. Examples for French and English include *a, an, on* and *par*.

The second matching predicate was just like the first, except that it also evaluated to TRUE whenever the input token pair appeared as an entry in a translation lexicon. The translation lexicon was automatically extracted from an MRBD [Co+91].

## 3.2 Point Selection

As illustrated in Figure 3, even short sequences of TPCs form characteristic patterns. In particular, TPCs have the following properties:

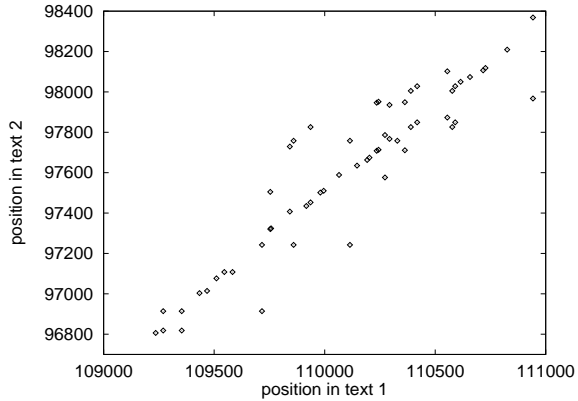

Figure 3: *Part of a typical scatterplot in bitext space. the true points of correspondence trace the true bitext map parallel to the main diagonal.*

- **Linearity**: TPCs tend to line up straight. Sets of points with a roughly linear arrangement are called **chains**.

- **Constant Slope**: The slope of a TPC chain is rarely much different from the bitext slope.

- **Injectivity**: No two points in a chain of TPCs can have the same x– or y–coordinates.

SIMR exploits these properties to decide which chains in the scatterplot might be TPC chains. The chain recognition heuristic involves two threshold parameters: **maximum point dispersal** and **maximum angle deviation**. Each threshold is used to filter candidate chains. First, the linearity of each chain is judged by measuring the root mean squared distance of the chain's points from the chain's least-squares line. If this distance exceeds the maximum point dispersal threshold, the chain is rejected. Second, the angle of each chain's least-squares line is compared to the arctangent of the bitext slope. If the difference exceeds the maximum angle deviation threshold, the chain is rejected. Lastly, chains that lack the injectivity property are rejected.

## 3.3 Reducing the Search Space

In a region of the scatterplot containing $n$ points, there are $2^n$ possible chains — too many to search by brute force. The properties of TPCs listed above provide two ways to constrain the search.

The Linearity property leads to a constraint on the chain size. Chains of only a few points are unreliable, because they often line up straight by coincidence. Chains that are too big will span too long a segment of the TBM to be well approximated by a line. SIMR chooses a fixed chain size $k$, $6 \leq k \leq 9$. Fixing the chain size at $k$ reduces the number of candidate chains to

$$\binom{n}{k} = \frac{n!}{(n-k)!k!}.$$

For typical values of $n$ and $k$, $\binom{n}{k}$ can still reach into the millions. The Constant Slope property suggests another constraint: SIMR should consider only chains that are roughly parallel to the main diagonal. Two lines are parallel if the perpendicular displacement between them is constant. So, if we want to find chains that are roughly parallel to the main diagonal, we should look for chains whose points all have roughly the same displacement[2] from the main diagonal. Points with similar displacement can be grouped together by sorting, as illustrated in Figure 4. Then, chains that are most parallel to the main diagonal will be contiguous subsequences of the sorted point sequence. In a region of the scatterplot containing $n$ points, there will be only $n - k + 1$ such subsequences of length $k$. Sorting the points by their displacement is the most computationally expensive step in the recognition process.

SIMR's chain recognition heuristic accepts non-monotonic chains. This is a desirable property, because even languages with similar syntax, like French and English, have well-known differences in word order. For example, English (adjective, noun) pairs usually correspond to French (noun, adjective) pairs. Such inversions result in chains that contain a pattern like points 5 and 9 in Figure 4. SIMR has no problem accepting the inverted points, unlike bitext mapping algorithms that try to minimize the distance between TPCs. To my knowledge, no other bitext mapping algorithm allows non-monotonic map segments.

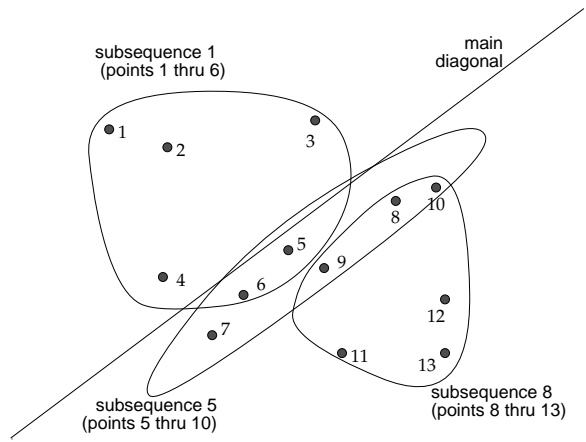

Figure 4: *The points of correspondence are numbered according to their displacement from the main diagonal. The chain most parallel to the main diagonal is always one of the contiguous subsequences of this ordering. For a fixed chain size of 6, there are $13 - 6 + 1 = 8$ contiguous subsequences in this region of 13 points. Of these 8, subsequence 5 is the best chain.*

You may wonder how SIMR will fare with languages that are less closely related, which have even more word order variation. This is an open question, but there is reason to be optimistic. To accommodate language pairs with vastly different word order, it may suffice for SIMR to increase the maximum point dispersal threshold, relaxing the linearity constraint on TPC chains.

## 3.4 Reducing Noise

The Injectivity property also leads to a heuristic which reduces the number of candidate chains, although the chief aim of this heuristic is to increase the signal-to-noise ratio in the scatterplot. The heuristic was introduced after inspection of several scatterplots in bitext spaces revealed a recurring noise pattern. This noise pattern is illustrated in Figure 5. It consists of correspondence points that line up in rows or columns associated with frequent token types. Token types like the English article "a" can produce one or more correspondence points for almost every sentence in the opposite text. Since only one of these correspondence points can be correct, all but one of the points in each row and column are noise. It's difficult to measure exactly how much noise is generated by frequent tokens, and of course the proportion is different for every bitext. Visual inspection of some scatterplots indicated that frequent tokens are often responsible for the lion's share of the noise. Reducing this source of noise makes it much easier for SIMR to stay on track.

---
[2] Displacement can be negative.

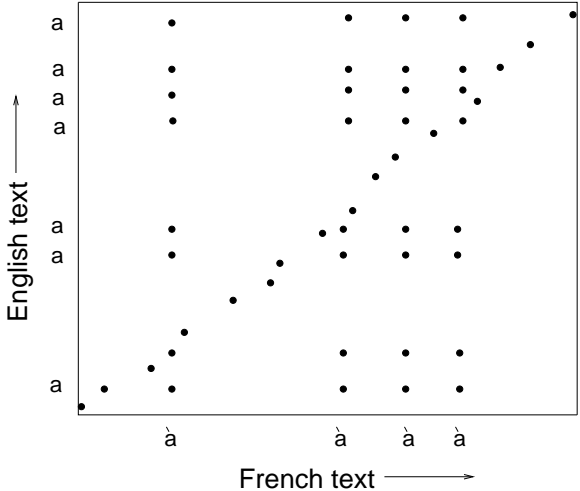

Figure 5: *Frequent token types cause false points of correspondence that line up in rows and columns.*

Other bitext mapping algorithms mitigate this source of noise either by assigning lower weights to correspondence points associated with frequent token types [Chu93] or by simply deleting frequent token types from the bitext [DCG93]. However, a frequent token type can be rare in some parts of the text. In those parts, the token type can provide valuable clues to correspondence. On the other hand, many tokens of a relatively rare type can be concentrated in a short segment of the text, resulting in many false correspondence points. The varying concentration of identical tokens suggests that more localized noise filters would be more effective. SIMR's localized search strategy provides the perfect vehicle for a localized noise filter.

The filter is based on another threshold parameter, the **maximum point ambiguity level (MaxPAL)**. For each point $p = (x, y)$, let X be the number of points in column $x$ within the search rectangle, and let Y be the number of points in row $y$ within the search rectangle. Then, ambiguity level of $p = X + Y - 2$.

Thus, if $p$ is the only point in its row and column, its ambiguity level is zero. SIMR ignores points whose ambiguity level exceeds the MaxPAL threshold. What makes this a localized filter is that only points within the search rectangle count towards each other's ambiguity level. This means that the ambiguity level of a given point can increase as the search rectangle expands; the set of points that SIMR ignores can change dynamically.

### 3.5 Interpolation

A bitext map can be derived from a set of correspondence points by linear interpolation. The only complication is that linear interpolation is not well-defined for non-monotonic sets of points. It would be incorrect to simply connect the dots left to right, because the resulting function may not be one-to-one. To interpolate injective bitext maps, non-monotonic segments must be encapsulated in Minimum Enclosing Rectangles (MERs), as shown in Figure 6. A unique bitext map can be interpolated by using the lower left and upper right corners of the MER, instead of using the non-monotonic correspondence points.

### 3.6 Enhancements

There are many possible enhancements to the algorithm outlined above. The following subsections describe but two of the more interesting extensions in the current implementation.

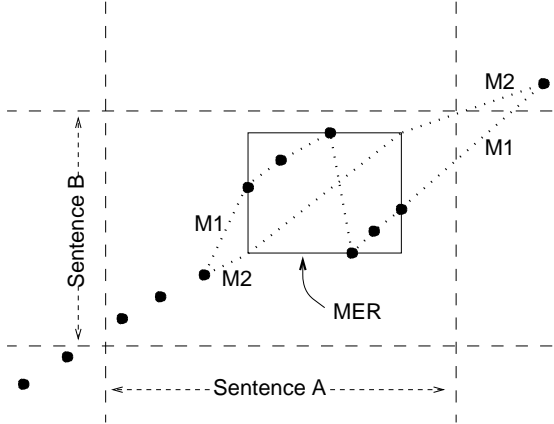
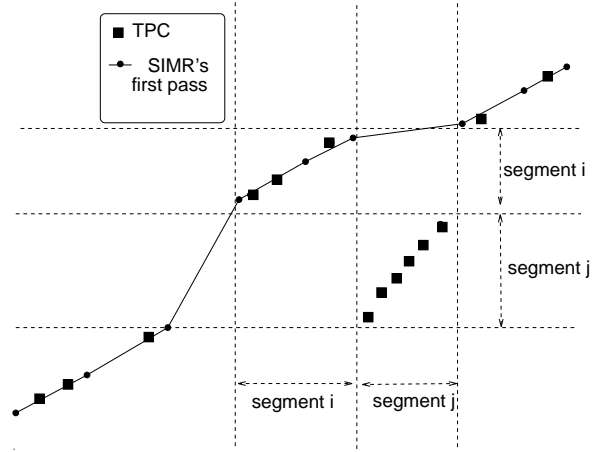

Figure 6: *Two text segments at the end of Sentence A were switched during translation, resulting in a non-monotonic segment. To interpolate injective bitext maps, non-monotonic segments must be encapsulated in Minimum Enclosing Rectangles (MERs). A unique bitext map can then be interpolated by using the lower left and upper right corners of the MER (map M2), instead of using the non-monotonic correspondence points (function M1).*

Figure 7: *Segments i and j switched placed during translation. If a more precise map is desired, these larger non-monotonic segments can be easily recovered during a second sweep through the bitext space. Any non-monotonic segment of the TBM will occupy the intersection of a vertical gap and a horizontal gap in the monotonic first-pass map.*

### 3.6.1 Large Non-monotonic Segments

SIMR has no problem with small non-monotonic segments inside chains. However, the expanding rectangle search strategy can miss larger non-monotonic segments, which cannot fit inside one chain. If a more precise map is desired, these larger non-monotonic segments can be easily recovered during a second sweep through the bitext space.

Non-monotonic TBM segments result in a characteristic map pattern, as a consequence of the injectivity of bitext maps. In Figure 7, the vertical range of segment j corresponds to a vertical gap in SIMR's first-pass map. The horizontal range of segment j corresponds to a horizontal gap in SIMR's first-pass map. Similarly, any non-monotonic segment of the TBM will occupy the intersection of a vertical gap and a horizontal gap in the monotonic first-pass map. Furthermore, switched segments are almost always adjacent and relatively short. Therefore, to recover non-monotonic segments of the TBM,
SIMR needs only to search gap intersections that are close to the first-pass map. There are usually very few such intersections that are also large enough to accommodate new chains, so the second-pass search requires only a small fraction of the computational effort of the first pass.

### 3.6.2 Local Slope Variation

To ensure that SIMR rejects spurious chains, the maximum angle deviation threshold must be set low. However, like any heuristic filter, this one will reject some perfectly valid candidates. The injectivity of bitext maps enables a method for recovering some of the rejected valid chains. Valid chains that are rejected by the angle deviation filter sometimes occur between two accepted chains, as shown in Figure 8. If chains C and D are accepted as valid, then the slope of the TBM between the end of Chain C and the start of Chain D must be much closer to the slope of Chain X than to the slope of the main diagonal. Chain X should be accepted. When SIMR makes its second-pass search for non-monotonic segments, it also searches for sandwiched chains in any space between two accepted chains that is large enough to accommodate another chain. This subspace of the bitext space will have its own main diagonal. The slope of this *local* main diagonal can be quite different from the slope of the *global* main diagonal.

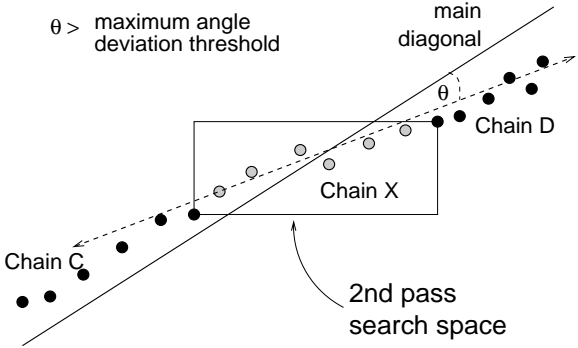

Figure 8: *Chain X is perfectly valid, even though it has a highly deviant slope. Such chains can be recovered by re-searching regions between accepted chains. The slope of the* local *main diagonal can be quite different from the slope of the* global *main diagonal.*

Another source of local slope variation is "non-linguistic" text, such as white space or tables of numbers. Usually, such text is copied "as is" during translation, resulting in regions of bitext space where the slope of the TBM is exactly 1. The problem is that these regions can be large enough to severely skew the slope of the main diagonal. Thus, they can fool SIMR into searching the whole bitext space for TPC chains whose slope is close to 1, even though most of the bitext map between "linguistic" parts of the bitext has a very different slope. Sometimes, the translation of non-linguistic text is completely erratic, especially where white space is concerned. Not surprisingly, SIMR cannot perform well on such text.

It should not be difficult to recognize bitext sections that consist of "non-linguistic" text. Then, SIMR will be better able to follow the variations in the slope of the TBM. This extension to SIMR is next in line.

### 3.7 Evaluation

The standard method of evaluating bitext mapping algorithms is to compare their output to a hand-constructed reference set of TPCs. Michel Simard of CITI graciously provided me with several such reference sets for French-English bitexts, including the same "easy" and "hard" Hansard bitexts that have been used to evaluate other bitext mapping and alignment algorithms in the literature [Chu93, SFI92, DCG93]. A non-Hansard reference set was used for SIMR's development. All of SIMR's parameters, namely the thresholds for maximum point dispersal, maximum angle deviation, maximum point ambiguity, and the LCSR used in the matching predicate, as well as the fixed chain size, were simultaneously optimized on this data set using simulated annealing [Vid93]. Different parameter settings considered by the optimization process resulted in different bitext maps for the development bitext. Each set of parameter values was scored according to the root mean squared error between the resulting bitext map and the reference set of TPCs. The best-scoring set of parameter values was used to evaluate SIMR.

SIMR was evaluated on the "easy" and "hard" Hansard bitexts. Note that these bitexts are so named because one was easier than the other for the alignment algorithm that was first evaluated on them. There is no *a priori* reason to believe that one or the other will be easier for SIMR. Table 1 compares SIMR's error distribution on these bitexts with that of the previous front-runner, `char_align`, as reported by Church [Chu93]. SIMR's RMS error is lower by more than a factor of 4. SIMR is also much more robust: it rarely errs by more than half the length of an average sentence. Such robustness has enabled at least one new commercial-quality application — automatic detection of omissions in translations [Mel96]. This task was impossible until now, because it cannot tolerate even a few wild errors, such as those produced by an independent implementation of `char_align` [Sim95].

Note that the error between a bitext map and each reference point can be defined as the hori-zontal distance, the vertical distance, or the distance perpendicular to the main diagonal. The latter distance will always be shortest, on average. Church [Chu93] did not specify which metric he used. Of the three possibilities, Table 1 conservatively reports the highest error estimates for SIMR. The lowest estimates for SIMR without the translation lexicon are an RMS error of 6.1 for the "easy" bitext and 5.4 for the "hard" bitext. With the translation lexicon, the lowest error estimates drop to 6.0 for the "easy" bitext and 4.6 for the "hard" bitext.

Table 1: *Comparison of error distributions for SIMR and* `char_align`, *in characters.*

| bitext | algorithm | median absolute error | 99th percentile | root mean squared error |
|---|---|---|---|---|
| "easy" | `char_align` | not reported | 200 | 57 |
| Hansard | SIMR | 0.49 | 50 | 13 |
| (7123 ref. pts.) | SIMR with MRBD | 0.61 | 49 | 13 |
| "hard" | `char_align` | 18 | 200 | 46 |
| Hansard | SIMR | 0.48 | 55 | 9.8 |
| (2693 ref. pts.) | SIMR with MRBD | 0.60 | 44 | 8.6 |

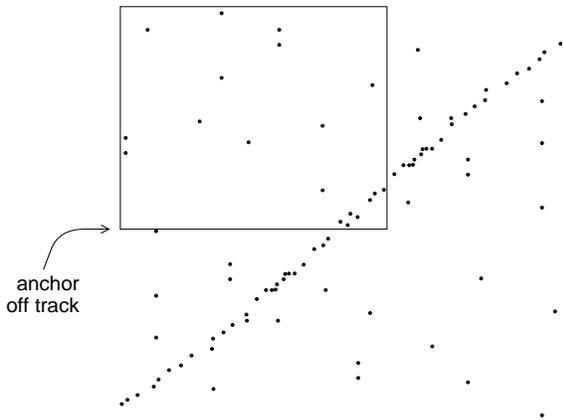

Figure 9: *TPCs are much more dense than false points of correspondence. This prevents SIMR from getting lost.*

### 3.8 Discussion

One concern about greedy algorithms is that if they wander off track, they may not be able to find their way back. Figure 9 shows how SIMR avoids this problem. The noise reduction heuristics described in Section 3.5 ensure that points of correspondence are very sparse, unless they are on the TBM trace. The expanding rectangle always finds its way back to the TBM before it finds a set of false points of correspondence that can fool the chain recognition heuristic.

The fixed chain size parameter plays an important role here. A larger set of false points of correspondence is less likely to take on a valid-looking arrangement. During optimization, SIMR occasionally veered off course when the fixed chain size was 5 or less. It rarely got lost with a fixed chain size of 6 and never with a fixed chain size of 7 or more. The optimal fixed chain size with respect to the RMS error metric was 9 when the translation lexicon was used, and 8 when it was not. The chances of 8 or 9 false points of correspondence satisfying the maximum point dispersal, maximum angle deviation, and maximum point ambiguity level thresholds are negligible. The development bitext used in the simulated annealing parameter optimization contained over 40000 words, so these conclusions can be made with confidence.

Finally, if SIMR does get lost, the resulting bitext map will contain telltale discontinuities. Such discontinuities can be automatically detected with high reliability [Mel96]. With this sanity check in place, manual verification should never be necessary.

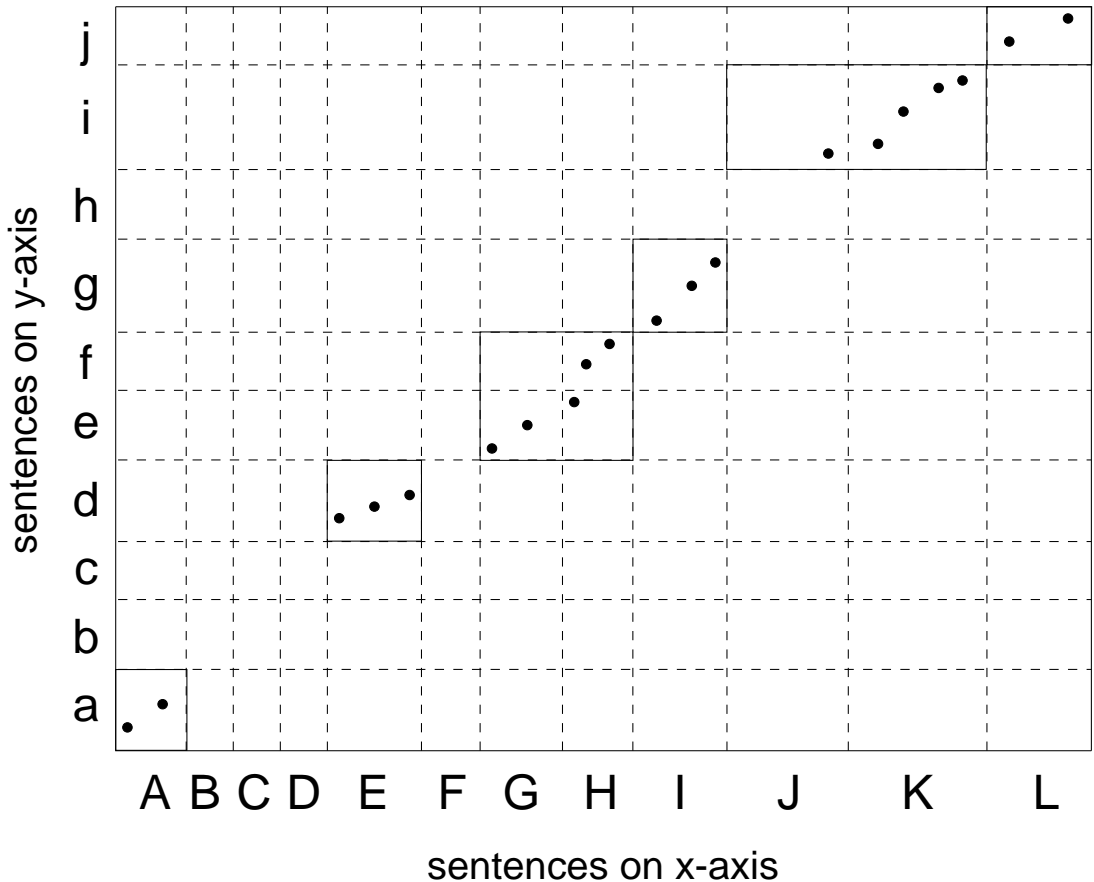

Figure 10: *Sentence boundaries form a grid over the bitext space. Each cell in the grid represents the product of two sentences, one from each component text. A point of correspondence inside cell (X, y) indicates that some token in sentence X corresponds with some token in sentence y; i.e. the sentences X and y correspond. So, for example, sentence E corresponds with sentence d. The aligned blocks are outlined with solid lines.*

## 4 Alignment

SIMR has no idea that words are often used to make sentences. It just outputs a series of corresponding token positions, leaving users free to draw their own conclusions about how the texts' larger units correspond. However, many existing translators' tools and machine translation strategies are based on aligned sentences. What can SIMR do for them?

There are several papers in the literature about bitext alignment. The algorithms that seem to work best rely on the high correlation between the lengths of corresponding sentences [BLM91, G&C91]. However, these algorithms can fumble in bitext sections that contain many sentences of very similar length, like this vote record:

| English | French |
|---|---|
| ⋮ | ⋮ |
| Mr. McInnis? | M. McInnis? |
| Yes. | Oui. |
| Mr. Saunders? | M. Saunders? |
| No. | Non. |
| Mr. Cossitt? | M. Cossitt? |
| Yes. | Oui. |
| ⋮ | ⋮ |

Source: [Che93]

The only way to ensure a correct alignment in such regions is to look at the words. For this reason, Chen [Che93] adds a statistical translation model to the Brown et al. alignment algorithm, and Wu [Wu94] adds a translation lexicon to the Gale & Church alignment algorithm.

A set of points of correspondence leads to alignment more directly than a translation model or a translation lexicon, because points of correspondence are a relation between token instances, not between token types. Moreover, a set of correspondence points, supplemented with sentence boundary information, expresses **sentence correspondence**, which is a richer representation than sentence alignment. Figure 10 illustrates how sentence boundaries form a grid over the bitext space[3]. Each cell in the grid represents the intersection of two sentences, one from each component text. A point of correspondence inside cell (X,y) indicates that some token in sentence X corresponds with some token in sentence y; i.e. sentences X and y correspond. Thus, Figure 10 indicates that sentence e corresponds with sentences G and H.

In contrast to a correspondence relation, "an **alignment** is a segmentation of the two texts such that the $n$th segment of one text is the translation of the $n$th segment of the other." [SFI92] For example, given the token correspondences in Figure 10, the segment $\langle G, H \rangle$ should be aligned with the segment $\langle e, f \rangle$. If sentences $\langle X_1, \ldots, X_n \rangle$ align with sentences $\langle y_1, \ldots, y_n \rangle$, then $(\langle X_1, \ldots, X_n \rangle, \langle y_1, \ldots, y_n \rangle)$ is an **aligned block**. In geometric terms, aligned blocks are rectangular regions of the bitext space, such that the sides of the rectangles coincide with sentence boundaries, and such that no two rectangles overlap either vertically or horizontally. The aligned blocks in Figure 10 are outlined with solid lines.

SIMR's initial output has more expressive power than the alignment that can be derived from it. One illustration of this difference is that sentence correspondence can express inversions, but sentence alignment cannot. Inversions occur surprisingly often in real bitexts, even for sentence-size text units. Figure 10 provides another illustration. If, instead of the point in cell (H,e), there was a point in cell (G,f), the correct alignment for that region would still be $(\langle G, H \rangle, \langle e, f \rangle)$. If there were points of correspondence in both (H,e) and (G,f), the correct alignment would still be the same. Yet, the three cases are clearly different. If a lexicographer wanted to see a word in sentence G in its bilingual context, it would be useful to know whether sentence f is relevant.

Converting from sentence correspondence to sentence alignment is of dubious practical value. Nevertheless, in order to facilitate comparison of the geometric approach with other alignment algorithms, I have designed the **Geometric Sentence Alignment (GSA)** algorithm to reduce sets of correspondence points to alignments. The algorithm's first step is to perform a transitive closure over the input correspondence relation. For instance, if the input contains (G,e), (H,e), and (H,f), then GSA adds the pairing (G,f). Next, GSA forces all segments to be contiguous: If sentence Y corresponds with sentences x and z, but not y, the pairing (Y,y) is added. In geometric terms, these two operations arrange all cells that contain points of correspondence into non-overlapping rectangles, while adding as few cells as possible. The result is an alignment relation.

A complete set of TPCs, together with appropriate boundary information, guarantees a perfect alignment. Alas, the points of correspondence postulated by SIMR are neither complete nor noise-free. Fortunately, the noise in SIMR's output causes alignment errors in very predictable ways. GSA employs a couple of backing-off heuristics to eliminate most of the errors.

SIMR makes errors of omission and errors of commission. Typical errors of commission are stray points of correspondence like the one in cell (H, e) in Figure 10. This point indicates that $\langle G, H \rangle$ and $\langle e, f \rangle$ should form a 2x2 aligned block, whereas the lengths of the component sentences suggest that a pair of 1x1 blocks is more likely. In a separate development bitext, I have found that SIMR is usually wrong in these cases. To combat such errors, GSA re-aligns any aligned block that is not 1x1, using the Gale & Church length-based alignment algorithm [G&C91, Sim95]. Whenever the component sentence lengths suggest a more fine-grained alignment, SIMR's output is not trusted.

---
[3] The techniques presented in this section can be applied equally well to paragraphs, lists of items, or any other text units for which boundary information is available.

Typical errors of omission are illustrated in Figure 10 by the complete absence of correspondence points between sentences $\langle B, C, D \rangle$ and $\langle b, c \rangle$. This block of sentences is sandwiched between aligned blocks. It is highly likely that at least some of these sentences are mutual translations, despite SIMR's failure to find any points of correspondence between them. Therefore, GSA treats all empty blocks just like aligned blocks. If an empty block is not 1x1, GSA re-aligns it using a length-based algorithm, just like it would re-align any other many-to-many aligned block.

The most difficult problem occurs when an error of omission occurs next to an error of commission, like in blocks $(\langle \rangle, \langle h \rangle)$ and $(\langle J, K \rangle, \langle i \rangle)$. If the point in cell (J,i) should really be in cell (J,h), re-alignment inside the erroneous blocks would not solve the problem. A naive solution is to merge these blocks and then to re-align them using a length-based method. Unfortunately, this kind of alignment pattern, i.e. 0x1 followed by 2x1, is surprisingly often correct. Length-based methods assign very low probabilities to such pattern sequences and usually get them wrong. Therefore, GSA also considers the confidence level with which the length-based alignment algorithm reports its re-alignment. If this confidence level is sufficiently high, GSA accepts the length-based re-alignment; otherwise, the alignment indicated by SIMR's points of correspondence is retained. The minimum confidence at which GSA trusts the length-based re-alignment is a GSA parameter, which has been optimized on a separate development bitext.

Due to the paucity of development resources at my disposal, GSA's backing-off heuristics are somewhat *ad hoc*. Even so, GSA performs at least as well as other alignment algorithms, and usually better. Table 2 compares SIMR's accuracy on the "easy" and "hard" reference bitexts with the accuracy of two other alignment algorithms, as reported by Simard et al. [SFI92]. The error metric counts one error for each aligned block in the reference alignment that is missing from the test alignment. I know of one other alignment algorithm with a published quantitative evaluation [Che93], but the error metric is not comparable to the one used here.

More important than GSA's current performance is GSA's potential performance. With a bigger development bitext, more effective backing-off heuristics can be developed. More precise input would also make a big difference: GSA's performance will improve whenever SIMR's performance improves.

Although GSA sometimes backs off to a quadratic-time alignment algorithm, in practice its running time is linear in the number of input sentences. The points of correspondence in SIMR's output are sufficiently dense and precise that GSA backs off only for very small aligned blocks. When the translation lexicon was used in SIMR's matching predicate, the largest aligned block that needed to be re-aligned in the "easy" and "hard" test bitexts was 5x5. Without the translation lexicon, the largest re-aligned block was 7x7. So, GSA's running time is $O(kn)$, where $n$ is the number of input sentences and $k$ is a small constant proportional to the size of the largest re-aligned block.

Admittedly, GSA is only useful when a good bitext map is available. In such cases, there are three reasons to favor GSA over other options for alignment: One, it is simply more accurate. Two, its running time is linear in the size of the bitext, faster than dynamic programming methods. Therefore, three, it is not necessary to partially pre-align large bitexts before input to GSA. In contrast, alignment algorithms that use dynamic programming are unacceptably slow on large inputs. Before such an algorithm can process a large bitext, the bitext must be segmented into a set of smaller bitexts. When a large bitext contains no clearly marked text units such as paragraphs or sections, the first-pass alignment must be done manually [G&C91, SFI92].

Table 2: *Comparison of alignment algorithms. One error is counted for each aligned block in the reference alignment that is missing from the test alignment.*

| bitext | algorithm | errors, given paragraph alignments | % | errors, not given paragraph alignments | % |
|---|---|---|---|---|---|
| "easy" Hansard ($n = 7123$) | Gale & Church (1991) | not available | | 128 | 1.8 |
| | Simard et al. (1992) | 114 | 1.6 | 171 | 2.4 |
| | SIMR/GSA | 104 | 1.5 | 115 | 1.6 |
| | SIMR/GSA with MRBD | 80 | 1.1 | 90 | 1.3 |
| "hard" Hansard ($n = 2693$) | Gale & Church (1991) | not available | | 80 | 3.0 |
| | Simard et al. (1992) | 50 | 1.9 | 102 | 3.8 |
| | SIMR/GSA | 50 | 1.9 | 61 | 2.3 |
| | SIMR/GSA with MRBD | 45 | 1.7 | 48 | 1.8 |

SIMR produced bitext maps for over 200 megabytes of the Canadian Hansards. GSA converted these maps into alignments. The Linguistic Data Consortium plans to publish both the maps and the alignments in the near future.

## 5 Conclusion

The Smooth Injective Map Recognizer (SIMR) has five advantages over previous bitext mapping algorithms. First, it lowers average errors by more than a factor of 4. Second, it avoids very large errors, improving robustness to a level that enables new commercial-quality applications. Third, it does not require large amounts of computer memory to run. Fourth, it accepts non-monotonic segments to account for inversions and word order differences. Fifth, its output can be converted quickly and easily into an accurate sentence alignment.

There are many possible extensions to this work. One interesting observation is that aligned sentences can be used to induce translation lexicons, and translation lexicons are an important information source for bitext mapping and alignment [K&R93, Che93]. I plan to explore an interactive loop between SIMR, GSA and my algorithm for inducing translation lexicons [Mel95].

It would also be interesting to experiment with SIMR and GSA on language pairs that are not as closely related as English and French. The only technique for mapping between more disparate languages that has been rigorously evaluated [Wu94] relies on length correlations sprinkled with some lexical information. From this point of view, Wu's technique is similar to the one used by Simard et al. [SFI92]. So, I am eager to see whether the geometric approach will compare as favorably to Wu's results on English and Chinese as it has to Simard et al.'s results on English and French.

## Acknowledgments

This research began while I was a visitor at the Centre d'Innovation en Technologies de l'Information in Laval, Canada. I am indebted to Pierre Isabelle for informing me that the bitext mapping problem is far from being solved. This paper has benefited tremendously from the insights and comments of the following people: Mike Collins, Jason Eisner, George Foster, Pierre Isabelle, Elliott Macklovitch, Mitch Marcus, Adwait Ratnaparkhi, Michel Simard, Eero Simoncelli, Matthew Stone, Lyle Ungar and three anonymous reviewers. My work was partially funded by ARO grant DAAL03-89-C0031 PRIME and by ARPA grants N00014-90-J-1863 and N6600194C 6043.